\documentclass[10pt,twocolumn,letterpaper]{article}
\usepackage{balance}
\usepackage{cvpr}
\usepackage{times}
\usepackage{epsfig}
\usepackage{graphicx}
\usepackage{amsmath}
\usepackage{amssymb}
\usepackage{subcaption}
\usepackage{booktabs, mathtools, algorithm,comment}
\usepackage[noend]{algpseudocode}
\usepackage[all]{nowidow}

\usepackage[pagebackref=true,breaklinks=true,letterpaper=true,colorlinks,bookmarks=false]{hyperref}
\DeclareMathOperator*{\argmin}{argmin} 

 \cvprfinalcopy 

\def\cvprPaperID{****} 

\ifcvprfinal\fi
\begin{document}

\title{Adversarial Light Projection Attacks on Face Recognition Systems:\\A Feasibility Study}

\author{Dinh-Luan Nguyen\textsuperscript{1,2}, Sunpreet S. Arora\textsuperscript{1}, Yuhang Wu\textsuperscript{1}, and Hao Yang\textsuperscript{1}\\
\textsuperscript{1}Visa Research, Palo Alto CA USA 94306\\
\textsuperscript{2}Michigan State University, East Lansing MI USA 48824\\
{\tt\small Corresponding author email: sunarora@visa.com}
}

\maketitle

\begin{abstract}
   Deep learning-based systems have been shown to be vulnerable to adversarial attacks in both digital and physical domains. While feasible, digital attacks have limited applicability in attacking deployed systems, including face recognition systems, where an adversary typically has access to the input and not the transmission channel. In such setting, physical attacks that directly provide a malicious input through the input channel pose a bigger threat. We investigate the feasibility of conducting real-time physical attacks on face recognition systems using adversarial light projections. A setup comprising a commercially available web camera and a projector is used to conduct the attack. The adversary uses a transformation-invariant adversarial pattern generation method to generate a digital adversarial pattern using one or more images of the target available to the adversary. The digital adversarial pattern is then projected onto the adversary's face in the physical domain to either impersonate a target (\textit{impersonation}) or evade recognition (\textit{obfuscation}). We conduct preliminary experiments using two open-source and one commercial face recognition system on a pool of 50 subjects. Our experimental results demonstrate the vulnerability of face recognition systems to light projection attacks in both white-box and black-box attack settings.
\end{abstract}

\section{Introduction}

Deep learning-based systems are typically designed under the assumption that the inputs/examples presented to the system during the test/operational phase follow the same underlying distribution as the examples used to train the system. However, recent research has shown security vulnerabilities of such systems when input test examples are intentionally crafted to cause the system to produce incorrect results (called \textit{adversarial examples}) ~\cite{szegedy2013intriguing, goodfellow2014explaining}. Most adversarial examples on convolutional neural network architectures (typically used in image classification scenarios including face recognition) are generated by perturbing the pixel intensities directly in the digital domain ~\cite{madry2017towards, moosavi2016deepfool, carlini2017towards}. These digital attacks, however, do not directly translate into the physical domain where an adversary has access to the open-camera channel. In such setting, the adversary usually does not have access to the image captured using the camera that is input to the convolutional neural network. Specifically, consider a face recognition system, that is deployed, such that it captures a face image of a subject and compares it to the enrolled faces to validate or establish the identity of the subject. While security mechanisms can be enforced to safeguard the digital storage and transmission of facial data captured using the camera, an adversary can potentially trick the system by providing a malicious input to the camera directly ~\cite{kurakin2016adversarial, eykholt2018robust}.

\begin{figure}[!tbp]
\centering
  \begin{subfigure}[b]{0.35\columnwidth}
    \centering
    \includegraphics[width=\columnwidth]{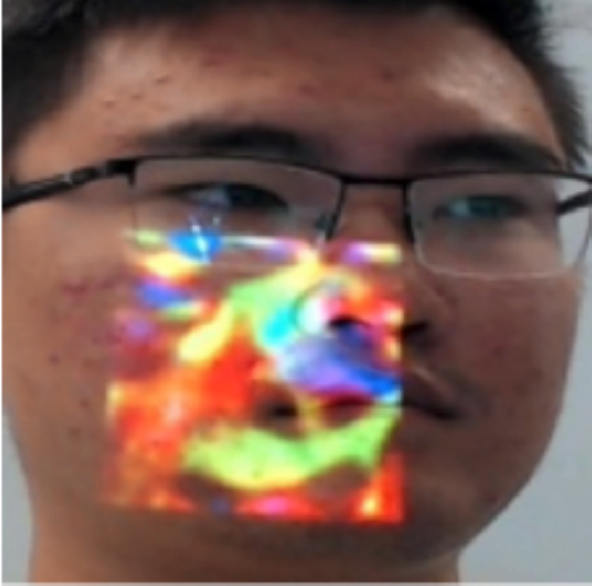}
    \caption{}
    \label{fig:1a}
  \end{subfigure}
  \hspace{18 pt}
  \begin{subfigure}[b]{0.35\columnwidth}
    \centering
    \includegraphics[width=\columnwidth]{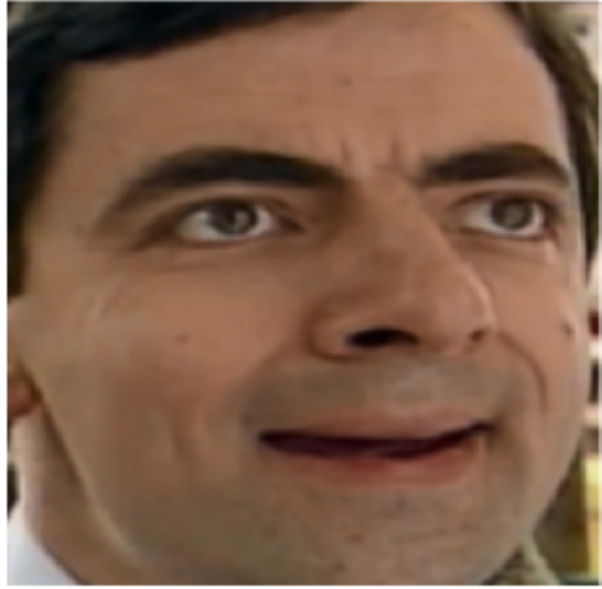}
    \caption{}
    \label{fig:1b}
  \end{subfigure}
\caption{Example of impersonation attack on FaceNet \cite{schroff2015facenet} in white-box setting. Shown in (a) is the captured image of the adversary's face with adversarial light projected in the physical domain that is recognized as the target (b).}
\label{fig:impersonation_whitebox}
\end{figure}

\begin{figure}[!tbp]
\centering
\includegraphics[width=\columnwidth]{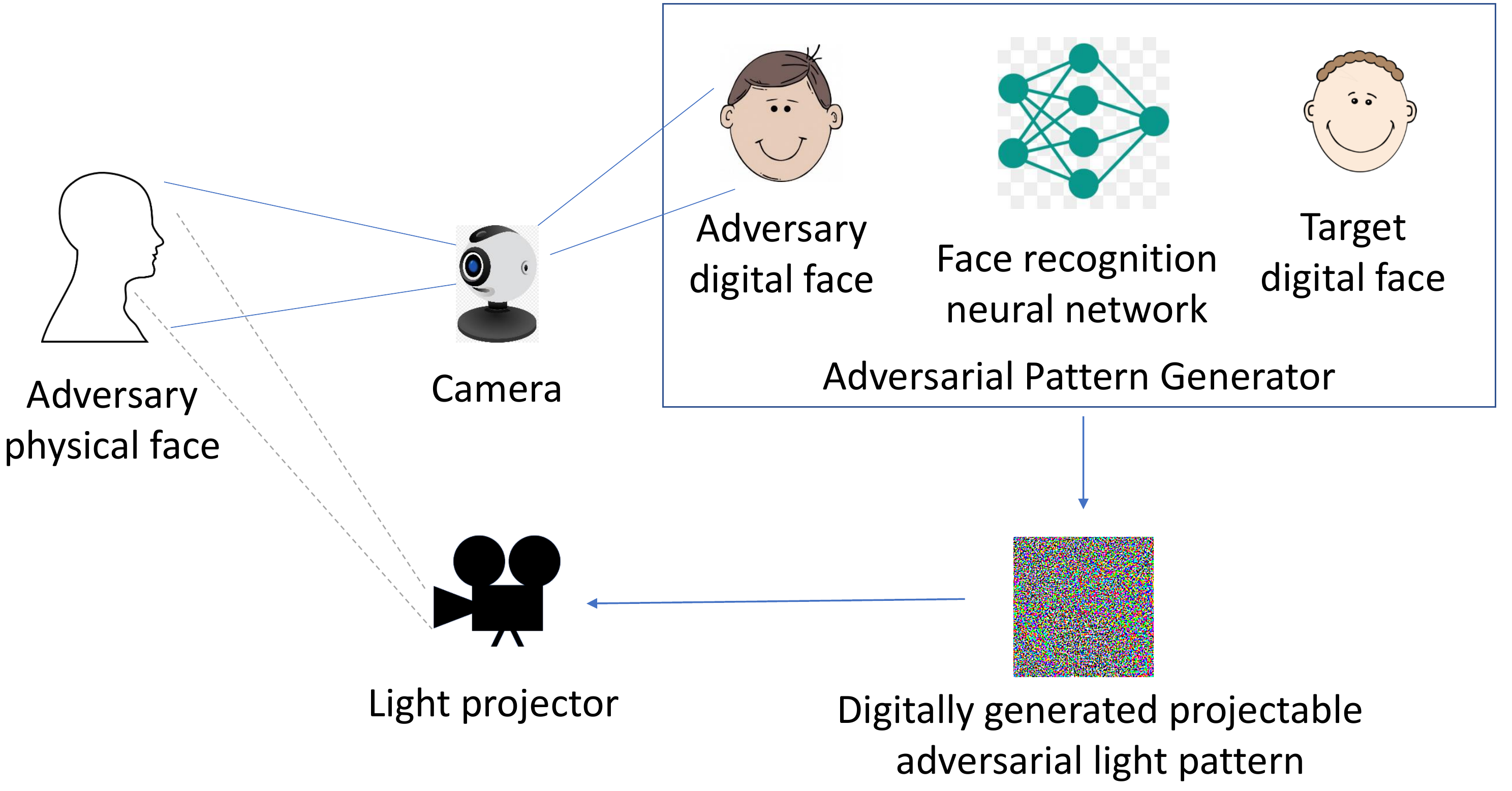}
\caption{Setup used for conducting real-time adversarial light projection attacks on face recognition systems. First, the adversary captures his/her facial image using a camera, and uses one or more images of the target to (i) calibrate the camera-projector setup based on the attack environment, and (ii) generate a digital adversarial pattern. Next, the adversary projects the digital pattern onto the adversary's face in the physical domain using a projector to either impersonate a target or evade recognition.}
\label{fig_system}
\end{figure}

A subclass of physical attacks on face recognition systems called \textit{presentation or spoofing} attacks achieve this by creating physical spoofs using one or more face images of the target (\eg, 2D-printed face photos, 3D masks) ~\cite{marcel2014handbook}. The same objective can also be achieved by crafting physical adversarial artifacts such as glasses that an adversary can wear to either evade recognition or mimic a target ~\cite{sharif2016accessorize}. However, fabrication of physical adversarial artifacts generally requires a manufacturing method (\eg, 2D or 3D printing). In addition, the utility of physical artifacts is limited in conducting at-scale physical attacks targeting multiple users of a face recognition system by the type of physical specimens that can be fabricated in typical resource-constrained settings.

We investigate the feasibility of conducting a real-time physical attack on face recognition systems using adversarial light projections that can be used for impersonating different enrolled users (called \textit{impersonation}), or evading recognition (called \textit{obfuscation}). The adversary first calibrates the camera-projector setup and then uses a transformation-invariant adversarial pattern generation method to generate adversarial patterns in the digital domain. These digital patterns are subsequently projected onto the adversary's face to conduct impersonation or obfuscation attack. We refer to this attack as \textit{adversarial light projection attack}. As an example, impersonation is the goal when an adversary intends to obtain access to a resource, \eg, personal device protected with a target's face. Obfuscation, on the other hand, is the goal of an adversary blacklisted by law enforcement agencies who wants to evade recognition in scenarios such as border crossing. 

A similar idea was recently proposed for fooling deep learning classifiers designed for image classification systems~\cite{nichols2018projecting}. However, the authors did not evaluate the utility of their method in the context of face recognition systems. Another recent work \cite{zhou2018invisible} fabricated a wearable cap with infrared LEDs to attack face recognition systems. Although this work is similar in terms of its objective, our method does not require a wearable artifact and thus offers an easier alternative using off-the-shelf camera-projector setup (\eg, a portable mini projector~\cite{pico1}) for conducting physical attacks on facial recognition systems. Preliminary experiments conducted on 50 subjects show the vulnerability of state-of-the-art face recognition systems to adversarial light projection attacks in both white-box and black-box attack settings.

\subsection{Contributions}
The major contributions of this work include:
\begin{itemize}

    \item Investigation of real-time adversarial light projection attacks using off-the-shelf camera-projector setup on state-of-the-art face recognition systems.
    
    \item An efficient transformation-invariant adversarial pattern generation method suitable for conducting real-time adversarial light projection attacks.
    
    \item Demonstration of vulnerability of state-of-the-art face recognition systems to adversarial light projection attacks in both white-box and black-box settings.
    
\end{itemize}

\section{Related Work}
Existing research on adversarial attacks can be broadly classified into two major categories: \textit{digital} and \textit{physical} attacks. Given one or more examples from source and target class, \textit{digital} attack methods generate adversarial pattern(s) in the digital domain such that the pattern(s) results in a source class example being misclassified as a target class example (\textit{called targeted attack}), or the source class example being incorrectly classified as an example from a different class (\textit{called untargeted attack}), typically with high confidence of being in the target class. \textit{Physical} attacks extend this notion into the physical domain by using specially crafted adversarial artifacts for targeted or untargeted attacks. Below we summarize the major research in the two categories and contrast existing physical attack methods from the method presented here.

\subsection{Digital Attacks}
One of the first digital attack methods proposed by Szegedy \etal in 2013 called L-BFGS~\cite{szegedy2013intriguing} formulates the goal of adversarial pattern generation as an optimization problem, and uses a box-constrained optimizer and linear search to find the optimal solution. In 2014, Goodfellow \etal~\cite{goodfellow2014explaining} proposed Fast Gradient Sign Method (FGSM), a single-step adversarial pattern generation method that uses gradients computed from neural network parameters for adversarial pattern generation. Following this, multiple extensions of FGSM were introduced \cite{kurakin2016adversarial, dong2018boosting, wu2018understanding, xie2019improving}.

Shi \etal~\cite{shi2019curls} combine gradient ascent and descent with binary search to find the adversarial pattern with the least $\ell_2$ norm. One of the most popular adversarial pattern generation methods Projected Gradient Descent (PGD)~\cite{madry2017towards} uses gradient projection space as a bound to generate $\ell_{\inf}$ adversarial patterns. Other prominent methods include DeepFool~\cite{moosavi2016deepfool} which was proposed for conducting untargeted attacks with $\ell_p$ norm, and models adversarial pattern generation as a linear approximation problem, and SparseFool~\cite{modas2019sparsefool} that aims to generate adversarial patterns by modifying a minimal number of pixels. Another popular method proposed by Carlini and Wagner~\cite{carlini2017towards} uses gradient descent with a custom loss function to minimize the $\ell_p$ norm during adversarial pattern generation.

\subsection{Physical Attacks}
Kurakin \etal~\cite{kurakin2016adversarial} printed 2D adversarial patches containing objects overlaid by adversarial patterns to attack deep networks trained for the object recognition task. Several other researchers directly printed 2D adversarial patterns which are then manually attached to physical objects to attack object detection and classification algorithms ~\cite{chen2018shapeshifter, song2018physical, eykholt2018robust, zhao2018practical}. Similarly, Thys \etal~\cite{thys2019fooling} printed 2D adversarial patches to circumvent pedestrian detection classifiers. Athalye \etal~\cite{athalye2017synthesizing} proposed a transformation-invariant adversarial pattern generation scheme called Expectation of Transformations (EOT) to fabricate 3D adversarial objects designed to fool object classifiers. More recently, Li \etal~\cite{li2019adversarial} printed adversarial dots on a 2D transparent paper to provide adversarial input via the camera to an object recognition system. 
Although the aforementioned methods succeed in achieving their stated objective, they usually require extensive calibration of each 2D or 3D-printed artifact before fabrication. In addition, they also require fabrication of physical artifacts. On the other hand, the camera-projector setup used in the method presented here can be calibrated once based on the attack environment, and then subsequently used for conducting multiple real-time attacks targeting different enrolled users of a face recognition system. 

Similar to this work, Nichols and Jasper~\cite{nichols2018projecting} used a camera-projector setup to generate 2D adversarial dot patterns that are then projected onto the physical scene to attack object recognition systems. However, they did not use the setup for conducting impersonation or obfuscation attacks on face recognition systems. Zhou \etal~\cite{zhou2018invisible}, on the other hand, fabricated a wearable cap with infrared LEDs to fool face recognition systems. Although this work is identical to the method presented here in terms of its objective, our method does not require creation of a wearable artifact and thus offers an easier alternative using off-the-shelf camera-projector setup for conducting physical attacks on facial recognition systems.

\section{Adversarial Light Projection Attack}
The proposed adversarial light projection attack is performed in two steps: the first step is to calibrate the camera-projector setup based on the attack environment and compute the adversarial pattern in the digital domain that can be used to either evade recognition or impersonate a target, and the second step is to project the computed digital adversarial pattern onto the adversary's face using the projector to attack the deployed face recognition system (see Figure ~\ref{fig_system}).

\subsection{Assumptions}
In the first step, the adversary is assumed to have either \textit{white-box} access\footnote{\textit In {white-box} setting, the adversary knows the internal details of the model including the architecture and trained weight parameters.} or \textit{black-box access}\footnote{In \textit{black-box} setting, the adversary only knows the decision/output of the model for one or more inputs.} to the deployed face recognition algorithm that the adversary intends to attack. White-box is a reasonable assumption for face recognition algorithms such as FaceNet \cite{schroff2015facenet} and SphereFace \cite{liu2017sphereface} that are available in open-source. On the other hand, commercial face recognition systems often only provide black-box access. So we assume that the adversary uses an open-source algorithm to generate adversarial patterns to attack the black-box system. This assumption exploits the property that adversarial patterns are highly transferable across deep network architectures.

Additionally, we assume that the adversary has access to an image of the target (in case of impersonation attack). 
Further, the adversary has access to a camera to capture the adversary's own face image in order to compute the adversarial light pattern, and a projector that is able to project light patterns on his/her own face in the physical domain in order to conduct the attack. In addition, the adversary is assumed to have either access to or reasonable prior knowledge of the environment where the face recognition system to be attacked is deployed. This is to ensure that the adversarial light pattern can be calibrated based on the attack environment before projection.

\begin{figure}[!tbp]
\centering
\includegraphics[width=\columnwidth]{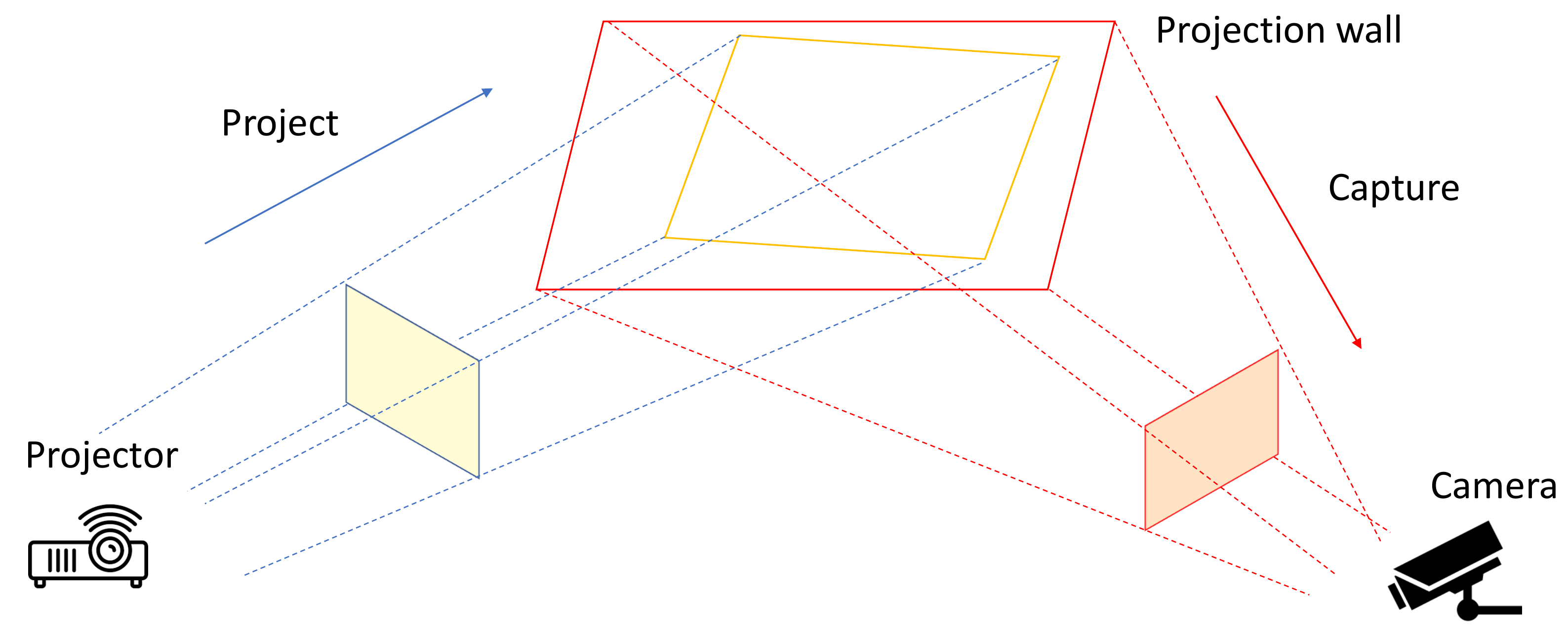}
\caption{Different viewpoints in the camera-projector setup. The physical adversary is assumed to be in the view of both the camera and the projector.}
\label{fig_camera_projector}
\end{figure}

\subsection{Practical Considerations}
\label{subsec:prac_cons}
Adversarial light projection attacks are inherently challenging because of their unconstrained nature. Below we discuss key practical considerations critical to the success of such attacks:
\begin{itemize}
    \item \emph{Environmental factors}, for example ambient and positional lighting, and their interplay with the projected light. Calibration of the attack setup based on the attack environment is therefore integral to the success of the attack (see Section \ref{section_calibration}).
    
    \item \emph{Intra-adversary facial variations} especially due to slight physical movements of the adversary, \eg, head movements and changes in the distance to the camera, while conducting the attack. Generation of adversarial patterns that are relatively invariant to such variations is therefore critical to the success of the attack (see Section \ref{sec_trans_invariant}).
    
    \item \emph{Intra-target facial variations} because the adversary would typically not have access to the enrolled images of the target in the deployed face recognition system. Instead, the adversary would have target images captured in a different context, such as social media. Hence it is important that the generated adversarial pattern be robust to the target's facial variations (see Section \ref{sec_trans_invariant}).
    
\end{itemize}

\section{Attack Setup Calibration}
\label{section_calibration}
There are two key calibration steps integral to success of the attack: (i) \emph{position calibration}: to ensure that the adversarial pattern generated in the digital domain can be projected onto the appropriate region of the adversary's face while conducting the attack, and (ii) \emph{color calibration}: to ensure that the digital adversarial pattern is reproduced with high fidelity by the projector as adversarial light.

\begin{figure}[!tbp]
\centering
\includegraphics[width=\columnwidth]{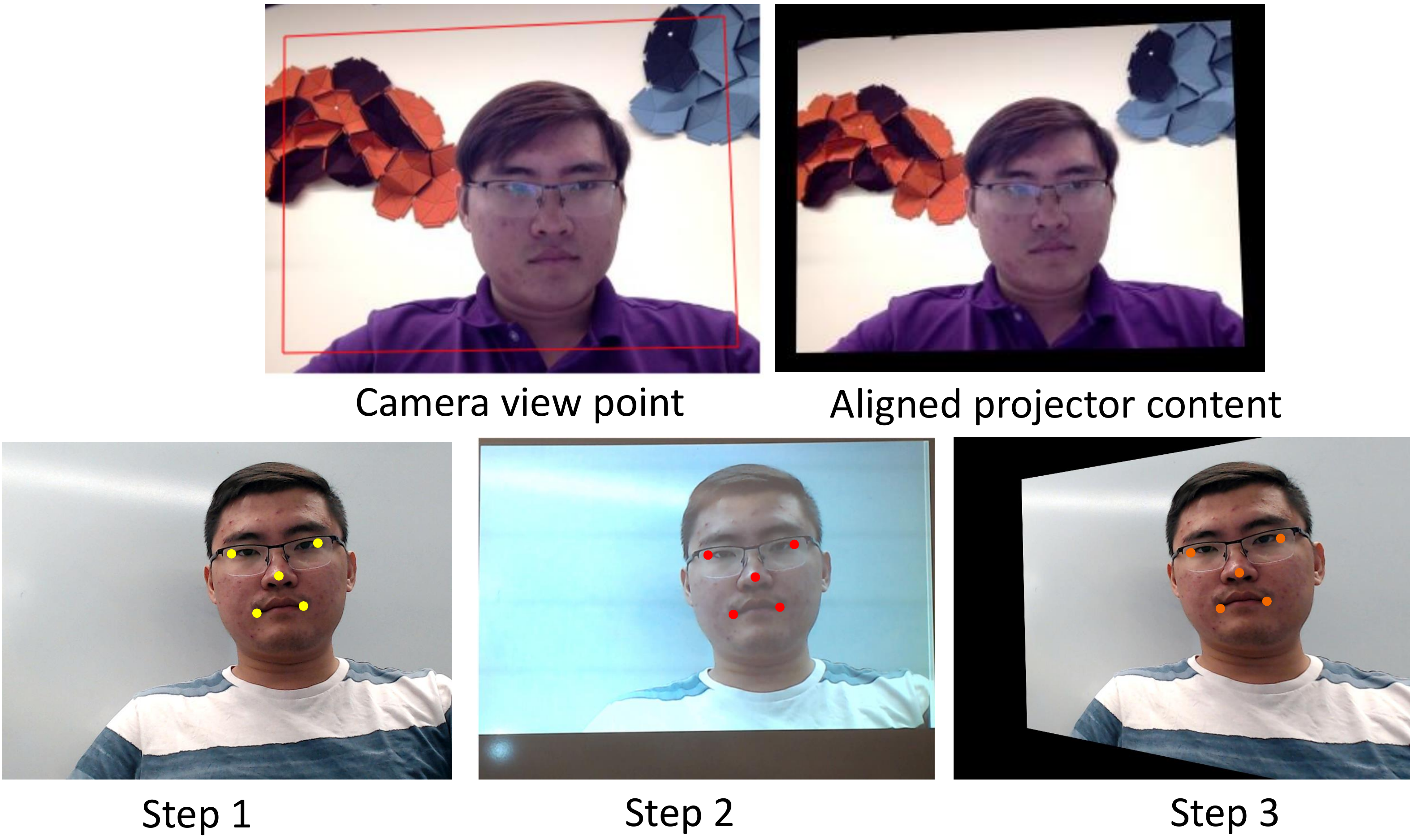}
\caption{Two possible methods for position calibration: (i) manual (top row) and (ii) automatic (bottom row).
\emph{Step 1}: Adversary's face is captured using the camera, and facial landmarks are either manually annotated or automatically detected; \emph{Step 2}: Projected scene (including the adversary’s face) is captured using the camera, and facial landmarks are manually annotated or automatically detected; \emph{Step 3}: Corresponding landmarks from step 1 and step 2 are used to compute the calibration matrix to calibrate the adversarial pattern before projection on the adversary's face.}
\label{fig_calibrate_position}
\end{figure}

\subsection{Position calibration}
Assume that the adversary is in view of both the camera and the projector (Figure \ref{fig_camera_projector}). There are two possible ways to perform position calibration: (i) \emph{manual}: the adversary manually annotates a small number (3-4) of corresponding points between the two views; or (ii) \emph{automatic}: a facial landmark detection algorithm is used to detect corresponding facial landmarks from the two views. Once the landmark correspondences are determined, a calibration matrix is computed to perform position calibration (see Figure \ref{fig_calibrate_position}).

\begin{algorithm*}[!t]
\caption{Transformation-Invariant Adversarial Pattern Generation}\label{algo_GOAT}
\textbf{Input:} Image of adversary $x$, target image $y$\\
\textbf{Output:} Adversarial pattern $x^{adv}$\\
\textbf{Notations:} $\mathcal{A}$: method to compute representative average image (equation \ref{eq_define_x_avg}); $f$: method to compute face embedding;\\ $\mathcal{M}$: distance metric (\eg, $\ell_p$, cosine); $clip$: method to clip intensity values (0-255); $F$: fusion function (\eg summation)$;\\ \alpha,\beta$: hyper-parameters
\begin{algorithmic}[1]
\Procedure{Optimize} {$x, y$}
\State $x^{temp}_0 \gets 0_{w \times h}$, 
$x^{avg}_0 \gets x$, 
and $x_0 \gets x$
\Comment{Initialization}
\While{not converge}
    \State $x^{avg}_{t-1} \gets \mathcal{A}(x^{avg}_{t-1})$
    \Comment{Compute representative average image}
    \State $x^{avg}_t \gets clip(x^{avg}_{t-1}+x^{temp}_{t-1}+\zeta_{\mathcal{N}(0,\sigma^2)})$ \Comment{Clipped representative image}
    \State $x_t \gets clip(x_{t-1}+x^{temp}_{t-1})$ \Comment{Clipped image of adversary}
    \State $\mathcal{D}^{avg} \gets \mathcal{M}(f(x^{avg}_t),f(y))$ and $\mathcal{D} \gets \mathcal{M}(f(x_t),f(y))$
    \Comment{Distance computation using metric $\mathcal{M}$}
    \State $\Delta \eta \gets \nabla_x F(\mathcal{D}^{avg}, \mathcal{D})|_{x=x^{temp}_{t-1}}$ \Comment{Compute update with respect to both $x$ and $x^{avg}$}
    \State $g_t \gets \beta \times g_{t-1} + \frac{\Delta \eta}{\|\Delta \eta \|_1}$ and $x^{temp}_t \gets x^{temp}_{t-1} + \alpha \times sign(g_t)$ \Comment{Gradient update with momentum}
\EndWhile
\State $x^{adv}=x_t-x$
\Comment{Adversarial pattern}
\EndProcedure
\end{algorithmic}
\end{algorithm*}

\subsection{Color calibration}
Let $\mathcal{C}$ and $\mathcal{G}$ be the color reproduction functions of the camera and projector, respectively, in the physical attack setting. The objective of color calibration is to find the color transformation function $\Upsilon$ given $\mathcal{C}$ and $\mathcal{G}$. Let $x=\mathcal{C}(x_0)$, where $x_0$ is the physical adversary, $x$ is the image of the adversary in the digital domain, and $h$ is the method used to generate adversarial pattern in the digital domain such as the one presented in section 5.3. Also, assume additive characteristic of the color reproduction functions, i.e. $\mathcal{C}(color1+color2)=\mathcal{C}(color1)+\mathcal{C}(color2)$. The relationship between the digital and physical domains can then be expressed as follows:

\begin{equation}
\label{eq_color}
    \begin{multlined}
        \mathcal{C}(x_0+\mathcal{G}(\Upsilon(h(x)))) = x + h(x)\\
        \Leftrightarrow \mathcal{C}(x_0)+\mathcal{C}(\mathcal{G}(\Upsilon(h(x))))) = x + h(x)\\
        \Leftrightarrow \mathcal{C}(\mathcal{G}(\Upsilon(h(x))))) = h(x)\\
        \Leftrightarrow \Upsilon = (\mathcal{C} \circ \mathcal{G})^{-1}
    \end{multlined}
\end{equation}

To empirically estimate the aforementioned relationship in the attack setting, (i) color calibration patterns are generated in the digital domain, (ii) the generated calibration patterns are projected on a white background in the physical domain, and (iii) the projected calibration patterns in the physical domain are captured in the digital domain using the camera. The color transformation function $\Upsilon$ is then computed by performing regression on the generated and camera-captured color pairs. In practice, we found that performing regression in $Lab$ color space results in more accurate color reproduction in the physical domain than standard $RGB$ color space.


\section{Transformation-Invariant Adversarial Pattern Generation}
\label{sec_trans_invariant}
Generation of adversarial patterns that are relatively invariant to intra-adversary facial variations is critical to the success of light projection attacks.
Let $x$ and $y$, respectively, be the images pertaining to the adversary and the target in the digital domain that are used in the adversarial pattern generation process. Existing methods (\eg, \cite{athalye2017synthesizing}, \cite{dong2019evading}) generate a transformation-invariant adversarial pattern $x^{adv}$ by applying different transformations on the adversary's image. In case of impersonation attack, the following optimization is solved:

\begin{equation}
\label{eq_naive}
    x^{adv} = \argmin_{\Delta x}{\sum_{i=0}^{k-1} {(w_i \mathcal{L}(\mathcal{T}_i(x) + \Delta x,y))}}, s.t. \|\Delta x\|_p \leq \varepsilon
\end{equation}

Here, ${T}_i$ corresponds to the $i_{th}$ transformation, $k$ is the total number of transformations, and $w_i$ corresponds to the weight of $T_i$ such that $\sum_{i=0}^{k-1}w_i=1$. Also, $\mathcal{L}$ is the loss function used in the adversarial pattern generation process. We assume that $\mathcal{L}$ is computed using a distance metric $\mathcal{M}$ in the face embedding space, and hence needs to be minimized. Alternatively, $\mathcal{L}$ could be computed using a similarity metric in which case it would need to be maximized. Equation \ref{eq_naive} involves computation of loss functions with respect to each transformation in each iteration. This is computationally intensive, and limits the application of these methods in the setting where an adversary wants to generate adversarial patterns in real-time.

\subsection{Computing representative adversary image}
Instead, our method computes an average representative image $x^{avg}$ of the adversary as follows:

\begin{equation}
\label{eq_define_x_avg}
    x^{avg}= \mathcal{A}(x) = w_0x+\sum_{i=1}^{k-1}(w_i\mathcal{T}_i(x)), s.t. \sum_{i=0}^{k-1}w_i=1
\end{equation}

For the impersonation task, the following optimization is then solved:

\begin{equation}
\label{eq_x_avg}
    x^{adv} = \argmin_{\Delta x}{\mathcal{L}(x^{avg}+\Delta x,y)}, s.t. \|\Delta x\|_p \leq \varepsilon
\end{equation}

Equation \ref{eq_x_avg} does not require explicit computation of loss functions for all possible transformations, yet explores a wide variety of transformation configurations in each iteration. The limitation though is that the optimization is performed with respect to a single representative image, and not with respect to expected loss pertaining to each transformation configuration. For real-time light projection attacks, this trade-off is desirable.

Also note that although equations \ref{eq_naive} and \ref{eq_x_avg} pertain to impersonation, equations for obfuscation can be formulated and solved in a similar manner.

\begin{figure*}[!tbp]
\centering
 \begin{subfigure}[b]{\textwidth}
    \centering
    \includegraphics[trim=110 100 50 150, clip, width=0.82\textwidth]{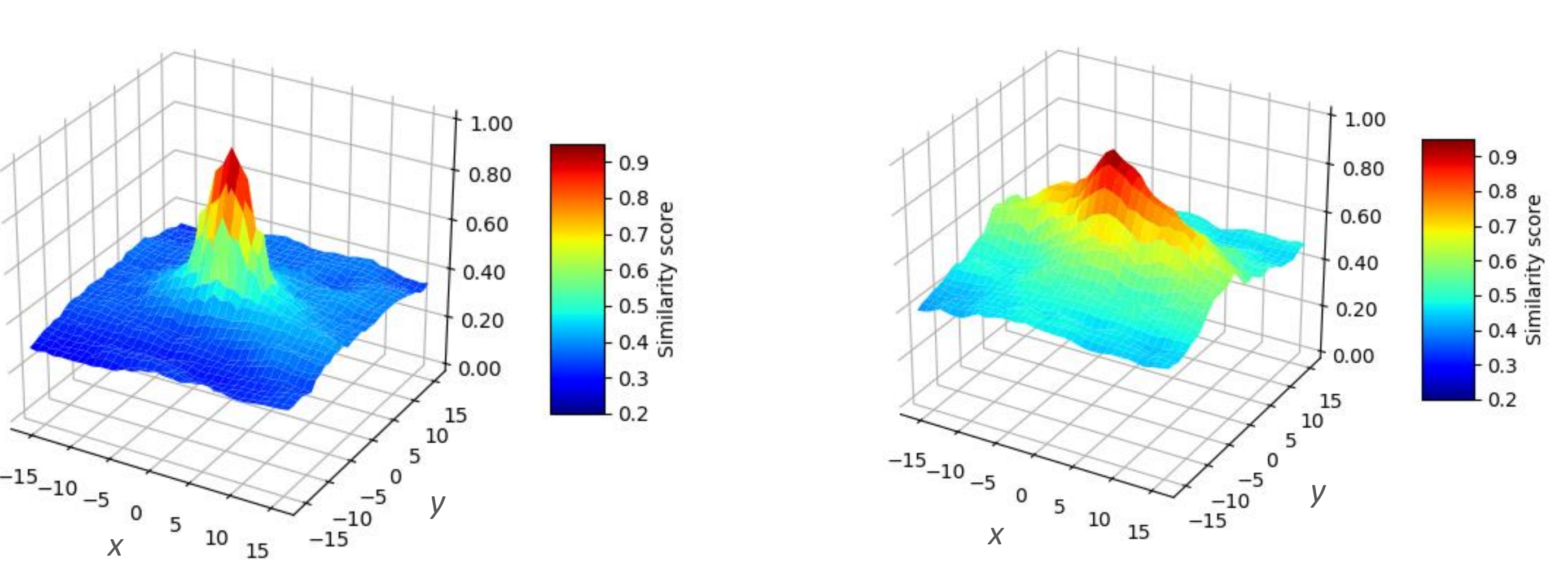}
    \caption{Translation}
    \label{fig:trans_a}
  \end{subfigure}
  \begin{subfigure}[b]{\textwidth}
    \centering
    \includegraphics[trim=35 50 40 100, clip, width=0.82\textwidth]{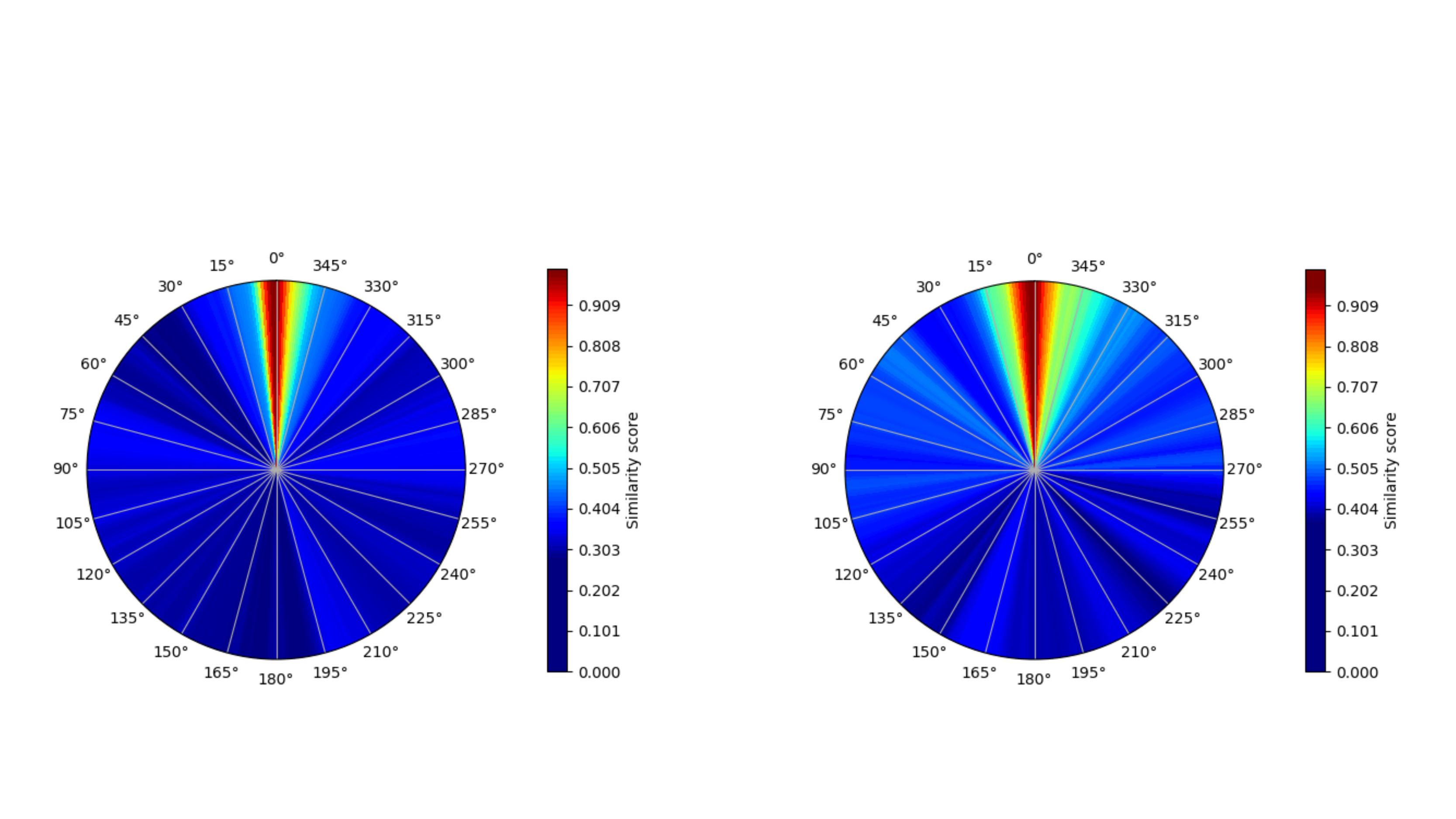}
    \caption{Rotation}
    \label{fig:trans_b}
  \end{subfigure}
\caption{Advantage of using the proposed transformation-invariant pattern generation method for (a) translation invariance and (b) rotation invariance. Left and right charts show, respectively, the similarity scores generated by FaceNet~\cite{schroff2015facenet} between an adversary's face image with projected adversarial light pattern and the target's face image, without and with (a) translation-invariance (in $x$ and $y$ direction in pixels) and (b) rotation-invariance (in degrees).}
\label{fig_transformation_exp}
\end{figure*}



\subsection{Using the original adversary image}
Optimizing with respect to the representative image $x^{avg}$ provides an efficient way to achieve transformation-invariance. However, to ensure that the generated pattern retains adversarial characteristics not only for the representative image $x^{avg}$ but for $x$ as well, optimization with respect to both $x^{avg}$ and $x$ is performed.
Figure \ref{fig_transformation_exp} illustrates example benefit for translation and rotation-invariance. Similar benefits were observed for other transformations.

\subsection{Proposed method}
Algorithm \ref{algo_GOAT} summarizes the proposed transformation-invariant pattern generation method. The method takes as input the image $x$ of the adversary and the image of the target $y$, and outputs the adversarial pattern $x^{adv}$.  The transformations used depend on the invariance objective (\eg, affine, perspective, photo-metric or others). The convergence criteria is either predefined number of steps or threshold on distance metric $M$ (step 7). A brightness term sampled from a normal distribution is used during each iterative update for obtaining invariance to slight illumination changes (step 5). Furthermore, a binary mask can be used to constrain the facial region for which the adversarial pattern is generated similar to \cite{sharif2016accessorize}.

\subsection{Using multiple images of target}
While the method described above focuses on intra-adversary invariance, it is desirable to impart invariance to intra-target variations as well to increase likelihood of success. For this, multiple images of the target can be used. Instead of a single target image, algorithm \ref{algo_GOAT} can then be optimized with respect to target embedding $y$ computed using multiple target images.

\begin{figure}[t]
\centering
  \begin{subfigure}[b]{0.38\columnwidth}
    \centering
    \includegraphics[trim={122 12 82 18},clip,width=\columnwidth]{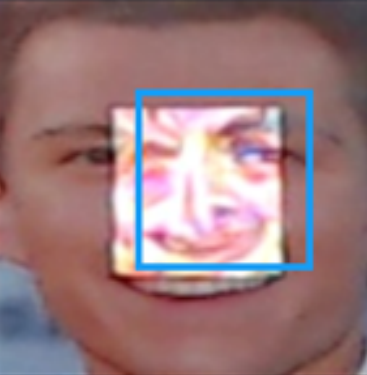}
    \caption{}
    \label{fig:imperson_a}
  \end{subfigure}\hspace{20pt}
  \begin{subfigure}[b]{0.39\columnwidth}
    \centering
    \includegraphics[width=\columnwidth]{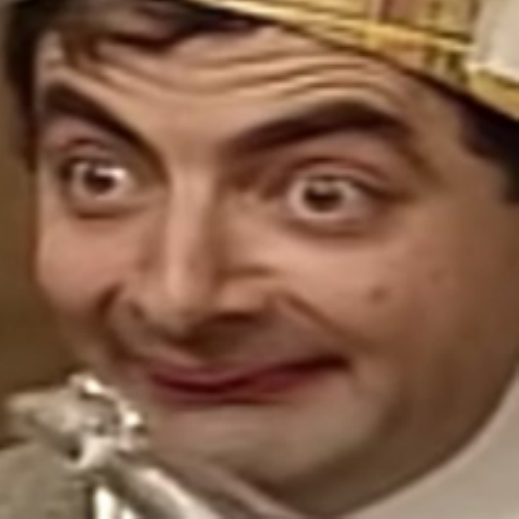}
    \caption{}
    \label{fig:imperson_b}
  \end{subfigure}
\caption{Example of impersonation attack on the commercial face recognition system in black-box setting. Shown in (a) is the captured image of the adversary's face with adversarial light projected in the physical domain that is recognized as target (b). This example is interesting because it appears that adversarial light resulted in failure of face detection (blue rectangle) yet the impersonation attack succeeded.}
\label{fig:impersonation_blackbox}
\end{figure}

\section{Experimental Evaluation}
To study the feasibility of adversarial light projection attacks, live subject experiments are performed with 50 subjects in total. Each experiment is conducted in a room with fixed lighting. A Logitech web camera and a Panasonic or Epson projector are used for the experiments.

\subsection{Experimental setup}
A web-based user interface is designed that takes real-time input of a subject's face using a web camera, and lets the subject select/upload face images of the target. The interface also lets the subject perform calibration based on the camera feed and the connected projector. Multi-task convolutional neural network (MTCNN)-based face detection and landmark estimation \cite{zhang2016joint} is used for automatic position calibration. For color calibration, the method described in section 4.2 is used. If necessary, the brightness and color intensity of the projector is manually tuned via the interface.

Post calibration, a python script executes the transformation-invariant pattern generation method (implemented in Tensorflow 2.0) for 100 iterations. Cosine is used as distance metric and multiplication as the fusion function. For gradient update, the parameters $\alpha$ and $\beta$ are set to $1$ and $0.7$ respectively. The following transformation configurations (assuming the origin is centered at midpoint of adversary's image) are considered during generation of the transformation-invariant pattern: $-40$ to $+40$ pixels translation in both $x$ and $y$ directions, $-\pi/3$ to $+\pi/3$ rotation, and $0.5-2$ times scaling. Each transformation configuration is assumed to be equally likely \ie the weight corresponding to each transformation is set to $\frac{1}{k}$ where $k=3$ is the number of transformations.

The computed digital pattern is projected using the projector in the form of adversarial light onto the subject's face. The subject's face with light projection is captured for about 30 seconds and used to attack a face recognition algorithm in real-time. The subject is instructed to make natural head movements (e.g. translation, rotation) during the duration of the attack. Similarity score between each captured image and the target face image is computed. If the computed score for any adversary-target image pair is above the threshold corresponding to False Accept Rate (FAR) of 0.01\%, the attack attempt is considered successful. FaceNet \cite{schroff2015facenet} and SphereFace \cite{liu2017sphereface} are the two face recognition algorithms used in white-box setting. In black-box setting, FaceNet is used to generate adversarial pattern to attack a commercial face recognition algorithm.

\begin{table}[t]
  \begin{center}
  \resizebox{\columnwidth}{!}{%
  \begin{tabular}{|l|c|c|c|}
      \toprule 
      \textbf{Experiment} & \textbf{\#Subjects} & \textbf{\#Attempts} & \textbf{Success Rate (\%)}\\
      \midrule 
      imp-fix-FN & 25 & 25 & 92.00\\
      imp-fix-SF & 25 & 25 & 84.00\\
      imp-fix-CO & 25 & 25 & 60.00\\
      imp-select-FN & 15 & 15 & 93.33\\
      imp-select-SF & 15 & 15 & 80.00\\
      imp-top5-FN & 10 & 50 & 88.00\\
      imp-top5-SF & 10 & 50 & 78.00\\
      \midrule
      obf-FN & 10 & 10 & 100.00\\
      obf-SF & 10 & 10 & 100.00\\
      obf-CO & 10 & 10 & 70.00\\
      \bottomrule 
    \end{tabular}}
    \caption{Impersonation (imp) and obfuscation (obf) experiments conducted on live subjects. FC, SF and CO refer to FaceNet, SphereFace and the commercial face recognition system respectively. Similarity score threshold corresponding to FAR=0.01\% is used to determine if the attack was successful.}
        \label{tab:table1}
  \end{center}
\end{table}

\subsection{Impersonation}
\begin{figure}[t]
\centering
  \begin{subfigure}[b]{0.52\columnwidth}
    \centering
    \includegraphics[trim={160 280 150 280},clip, width=\columnwidth]{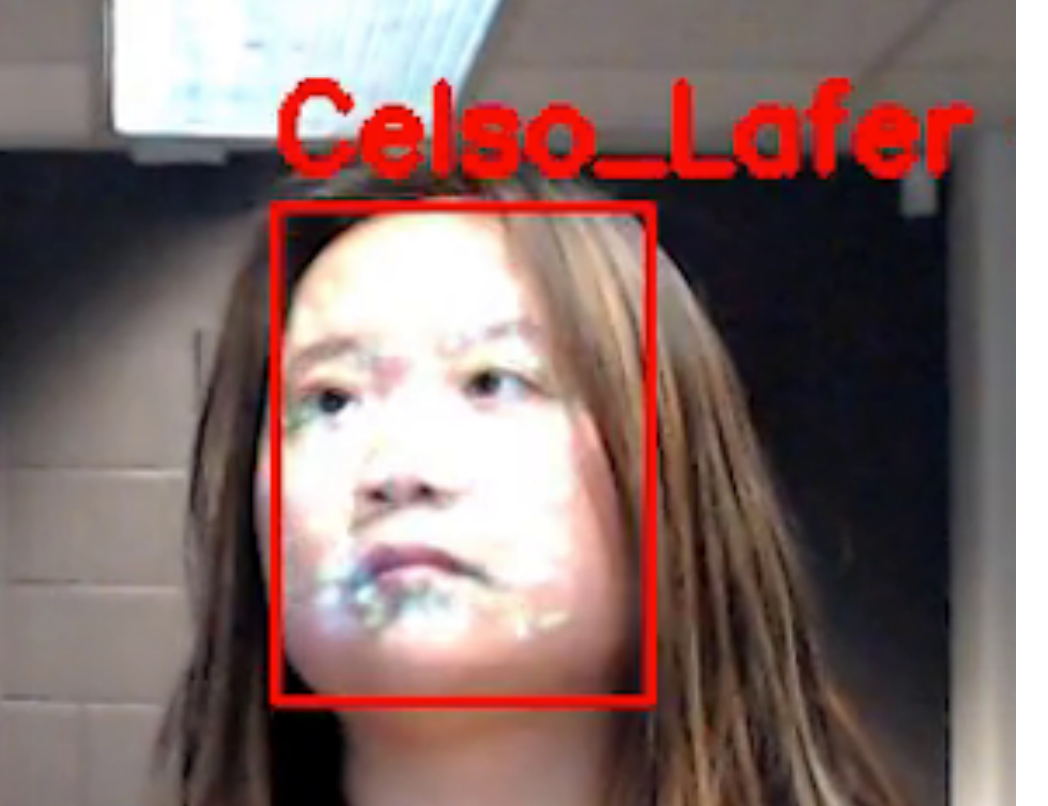}
    \caption{}
    \label{fig:2a}
  \end{subfigure}
  \hspace{5 pt}
  \begin{subfigure}[b]{0.37\columnwidth}
    \centering
    \includegraphics[trim={250 320 235 335},clip, width=\columnwidth]{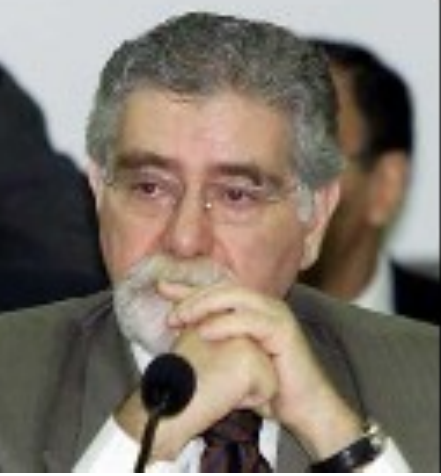}
    \caption{}
    \label{fig:2b}
  \end{subfigure}
\caption{Example of impersonation attack on SphereFace \cite{liu2017sphereface} in white-box setting. Shown in (a) is the captured image of the adversary's face with adversarial light projected in the physical domain that is recognized as target (b).}
\label{fig:impersonation_whitebox_2}
\end{figure}

For impersonation, a face image of a subject (adversary) captured using the camera, and a face image of the target (obtained from the web or a database) are used to generate the digital adversarial pattern. A different face image of the target (also obtained from the web or a database) is assumed to be enrolled in the face recognition system to be attacked. Impersonation attempts are made using different subject pools in the following scenarios.
\begin{itemize}
    \item \emph{Fixed target (imp-fix)}: Impersonating a fixed high-profile target (Rowan Atkinson). 25 subjects in total attempted this in both white-box and black-box setting. 23 and 21 attempts out of 25 on FaceNet and SphereFace, respectively, succeeded. In black-box setting, 15 out of 25 attempts on the commercial system succeeded.
    \item \emph{Selected target (imp-select)}: Impersonating any one of the given targets (Taylor Swift, Michael Phelps, or Albert Einstein) at random. A total of 15 subjects attempted this in white-box setting. 14 and 12 attempts out of 15 on FaceNet and SphereFace, respectively, succeeded.
    \item \emph{Top-5 similar targets (imp-top5)}: Given a database of target images (Labelled Faces in the Wild (LFW) database \cite{huang2008labeled}), impersonate top-5 most similar targets based on a face recognition algorithm. 10 different subjects attempted this on five most similar targets from LFW in white-box setting. 44 and 39 out of 50 attempts on FaceNet and SphereFace, respectively, succeeded.
\end{itemize}
\noindent Table \ref{tab:table1} summarizes the experimental results. Figures \ref{fig:impersonation_whitebox}, \ref{fig:impersonation_whitebox_2} and \ref{fig:impersonation_blackbox} show successful examples of impersonation, respectively, on FaceNet, SphereFace and the commercial face recognition system.

\subsection{Obfuscation}
\begin{figure}[t]
\centering
\includegraphics[trim={117 24 90 20},clip,width=0.45\columnwidth]{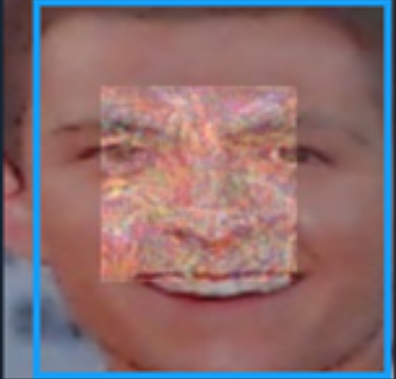}
\caption{Example of obfuscation attack on the commercial face recognition system in black-box setting. Adversarial light projected on the adversary's face in the physical domain results in successful obfuscation.}
\label{fig:obfuscation_blackbox}
\end{figure}

For obfuscation (obf), two face images of a subject (adversary-target pair) captured using the camera are used to generate digital adversarial pattern. A different face image of the same subject (target) is assumed to be enrolled. 10 different subjects attempted obfuscation attacks in both white-box and black-box setting. All 10 obfuscation attempts on FaceNet and SphereFace succeeded in white-box setting, whereas in black-box setting 7 out of 10 attempts on the commercial face recognition system succeeded. Table \ref{tab:table1} summarizes the experimental results. Figure \ref{fig:obfuscation_blackbox} shows a successful example of obfuscation on the commercial face recognition system.

\subsection{Failure Cases}
While most impersonation and obfuscation attempts are successful, failure of adversarial light projection attacks is observed due to one or more of the following reasons:
\begin{itemize}
    \item Light projection either covering the entire face or significantly occluding majority of the face resulting in failure of face detection. Projection of adversarial light on a particular part of the face, \eg, cheeks or forehead, is found to result in higher likelihood of success in practice.
    \item Strong ambient or directional lighting that overpowers the projected light. 
    \item Extreme facial pose of the adversary. In practice, however, this is less likely as the adversary is cooperative.
    \item Out of focus light projection on the adversary's face when the projector lens is not tuned appropriately. Manual tuning of the projector lens to ensure the projected light is properly focused on the adversary's face is important for a successful attack attempt in practice.
\end{itemize}
We plan to investigate failure cases in a more systematic manner in a follow-up study.

\section{Conclusions and Future Work}
We show the feasibility of conducting impersonation and obfuscation attacks using adversarial light projections on two open-source face recognition systems in white-box setting and a commercial face recognition system in black-box setting. Furthermore, an efficient transformation-invariant adversarial pattern generation method that enables an adversary to conduct light projection attacks in real-time is presented.

While we have shown the feasibility of light projection attacks, we have not systematically tested the likelihood of success at different operating thresholds of face recognition systems in different environments. One of our immediate goals, therefore, is to systematically investigate the impact of environmental and subject-dependent covariates (such as lighting, subject orientation and pose) on the repeatability of light projection attacks. Furthermore, we suspect that presentation attack detection methods designed for static attacks using 2D or 3D fabricated artifacts will be inadequate in defending against dynamic adversarial attacks such as light projection attacks. Therefore, we also plan to conduct an evaluation of existing defense mechanisms and develop novel defense mechanisms for such dynamic attacks.

\balance
{\small
\bibliographystyle{ieee_fullname}
\bibliography{CVPR20_bib}
}

\end{document}